\documentclass[a4paper,conference]{IEEEtran}
\usepackage{cite}

\ifCLASSINFOpdf
\else
\fi
\usepackage[dvipdfmx]{graphicx}
\usepackage[dvipdfmx]{color}

\usepackage{amsmath}

\usepackage{here}

\usepackage{subfigure}

\usepackage{multirow, array}
\usepackage{bbding}
\usepackage{amssymb}

\newcommand{\bhline}{\noalign{\hrule height 1.5pt}} 

\usepackage{xcolor}

\begin{document}
%
\title{Distortion-Adaptive Grape Bunch Counting for Omnidirectional Images}

\author{\IEEEauthorblockN{Ryota Akai\IEEEauthorrefmark{1}, Yuzuko Utsumi\IEEEauthorrefmark{1}, Yuka Miwa\IEEEauthorrefmark{2}, Masakazu Iwamura\IEEEauthorrefmark{1} and Koichi Kise\IEEEauthorrefmark{1}}
\IEEEauthorblockA{\IEEEauthorrefmark{1}Graduate School of Engineering, Osaka Prefecture University\\
Sakai, Osaka 591--8531, Japan\\
Email: akai@m.cs.osakafu-u.ac.jp, \{yuzuko, masa, kise\}@cs.osakafu-u.ac.jp}

\IEEEauthorblockA{\IEEEauthorrefmark{2}Research Institute of Environment, Agriculture and Fisheries, Osaka Prefecture\\
Habikino, Osaka 583--0862, Japan\\
Email: MiwaY@mbox.kannousuiken-osaka.or.jp}
}

%


\maketitle

\begin{abstract}
This paper proposes the first object counting method for omnidirectional images.
Because conventional object counting methods cannot handle the distortion of omnidirectional images, 
we propose to process them using stereographic projection, which enables conventional methods to obtain a good approximation of the density function.
However, the images obtained by stereographic projection are still distorted.
Hence, to manage this distortion, we propose two methods.
One is a new data augmentation method designed for the stereographic projection of omnidirectional images.
The other is a distortion-adaptive Gaussian kernel that generates a density map ground truth while taking into account the distortion of stereographic projection.
Using the counting of grape bunches as a case study, we constructed an original grape-bunch image dataset consisting of omnidirectional images and conducted experiments to evaluate the proposed method.
The results show that the proposed method performs better than a direct application of the conventional method, improving mean absolute error by 14.7\% and mean squared error by 10.5\%.

\end{abstract}

%
\IEEEpeerreviewmaketitle
\section{Introduction}\label{sec:intro}
%
%
Grapes are one of the most important crops in the world.
One essential task of growing grapes is bunch pruning, which reduces the number of immature green bunches on a grape trellis.
Grapevines, like other fruit trees, tend to bear a large number of fruit, but the amount of sugar that the plant can produce by photosynthesis is limited.
Therefore, pruning bunches is necessary to increase the sugar content to the level that is needed to sell the crop.
Because the photosynthetic capacity of a grapevine depends on the area of leaf that receives sunlight, the pruning standard is based on the number of bunches per unit area of trellis.
Usually, farmers divide the trellis into squares of a particular size and prune bunches so that the number of bunches in a square meets the standard.
To ensure this standard is met, farmers must count the bunches. 
Therefore, counting the bunches is a laborious but vital task in the process of bunch pruning.

\begin{figure}[tb]
\includegraphics[width=0.48\textwidth]{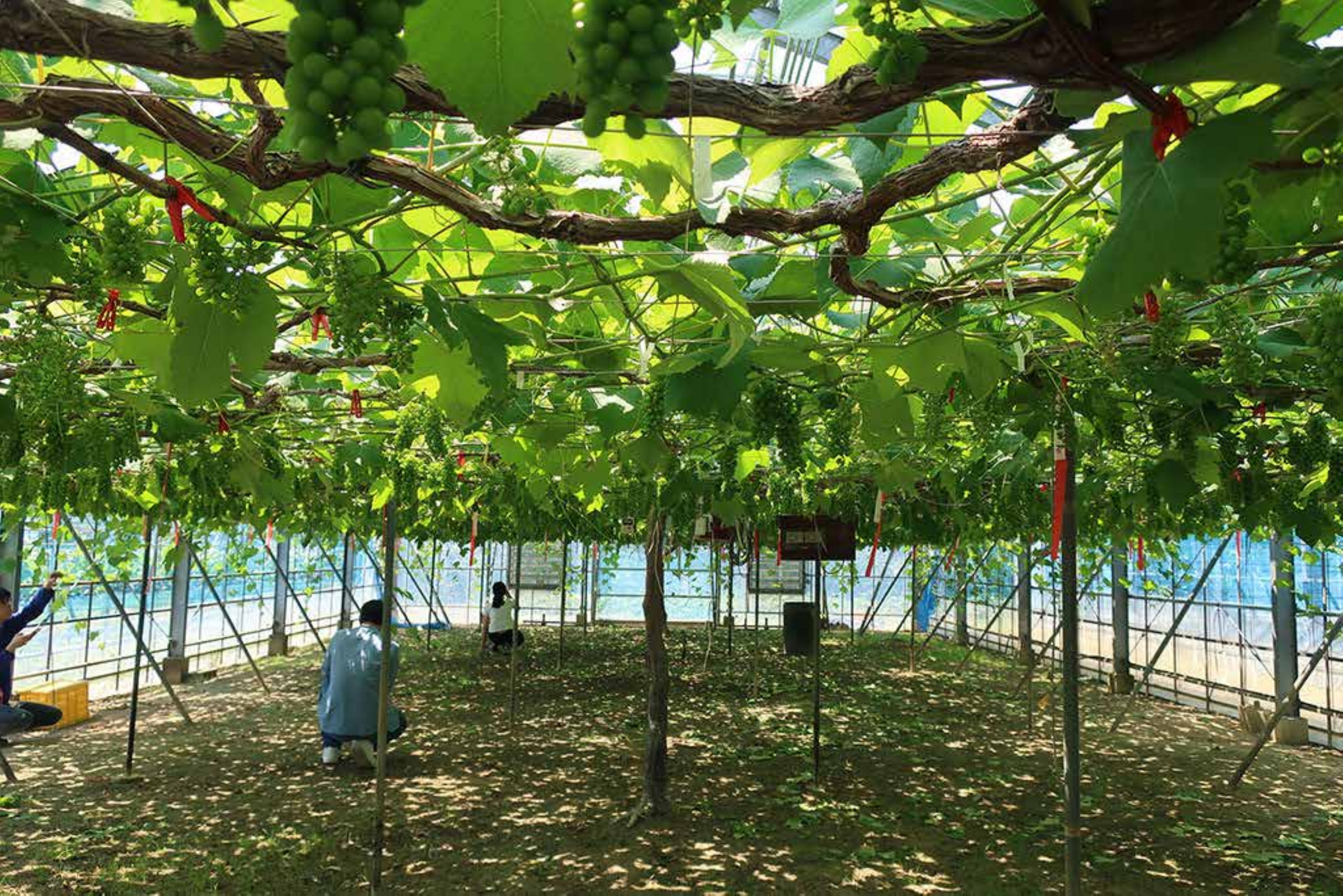}
\label{fig:trellis_perspective}
\caption{Side view of grapes on the trellis taken with a perspective camera. Red ribbons indicate the corners of a unit area.}
\end{figure}
To reduce the effort needed to manually count bunches, we automate this task using an object counting technique such as \cite{liCSRNetDilatedConvolutional2018,liuDecideNetCountingVarying2018,maBayesianLossCrowd2019,wanAdaptiveDensityMap2019,xiongOpenSetClosed2019, xieMicroscopyCellCounting2016,heAutomaticMicroscopicCell2019, guerrero-gomez-olmedoExtremelyOverlappingVehicle2015, mundhenkLargeContextualDataset2016, amatoCountingVehiclesDeep2019, Aich2017,Dobrescu2017,Giuffrida2018,Ubbens2018, Lempitsky2010}.
We describe how the farmer performs this task below.
\figurename~\ref{fig:trellis_perspective} shows grapes grown on a trellis.
The trellis is parallel to the ground, and bunches hang from it.
Hence, a natural way to capture images of the bunches is to take the images from under the trellis.
Usually, the height of the trellis is about 1.6 m, and the size of a unit area is 2 m $\times$ 2 m. 
Hence, to capture the whole region of a unit area, a camera with a wide field of view such as an omnidirectional camera is required.

\begin{figure*}[tb]
 \centering
 \subfigure[Equirectangular projection]{
 \includegraphics[width=0.48\textwidth]{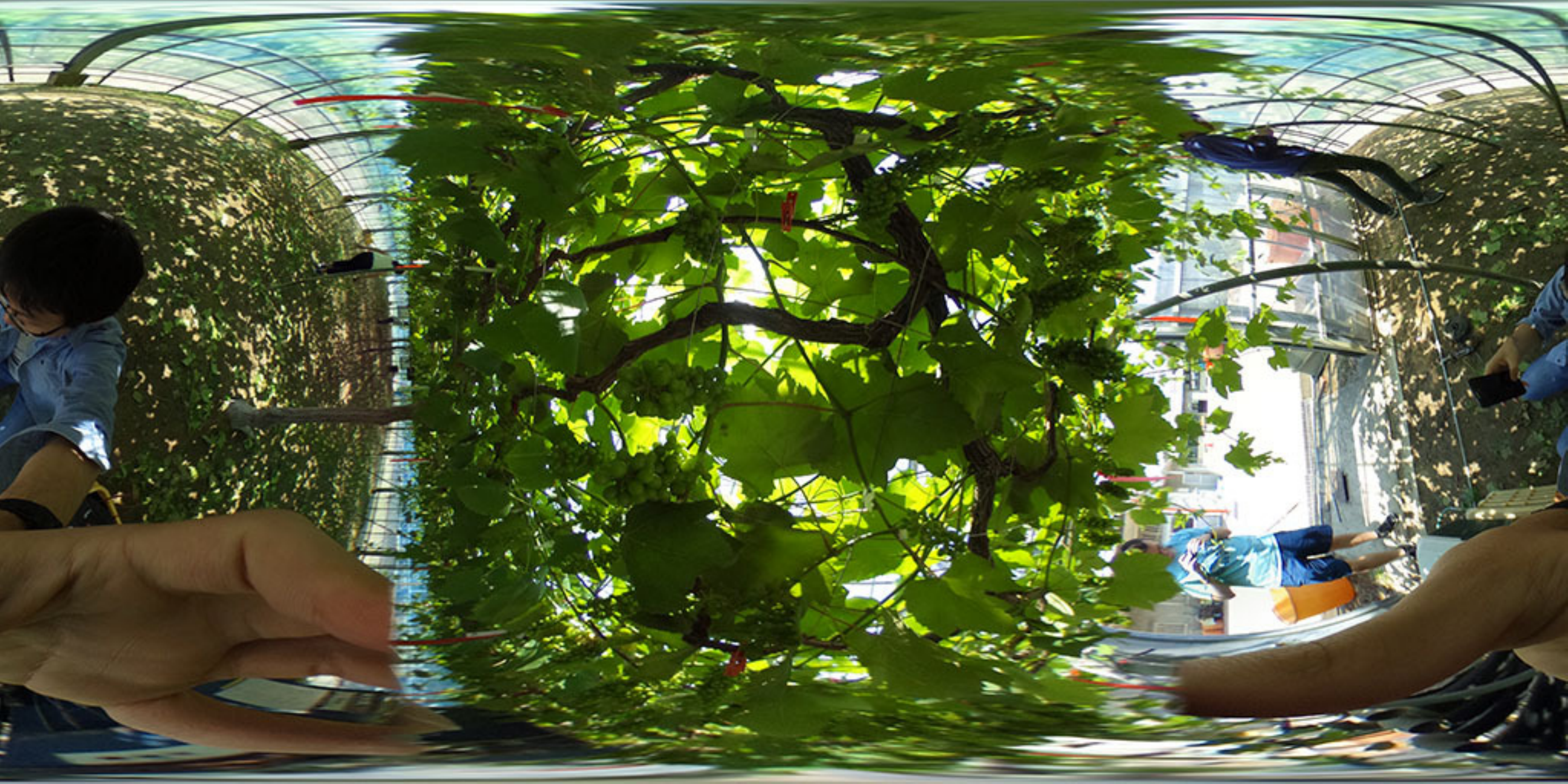}
 \label{fig:trellis_eq}
 }
 \subfigure[Stereographic projection]{
  \includegraphics[width=0.48\textwidth]{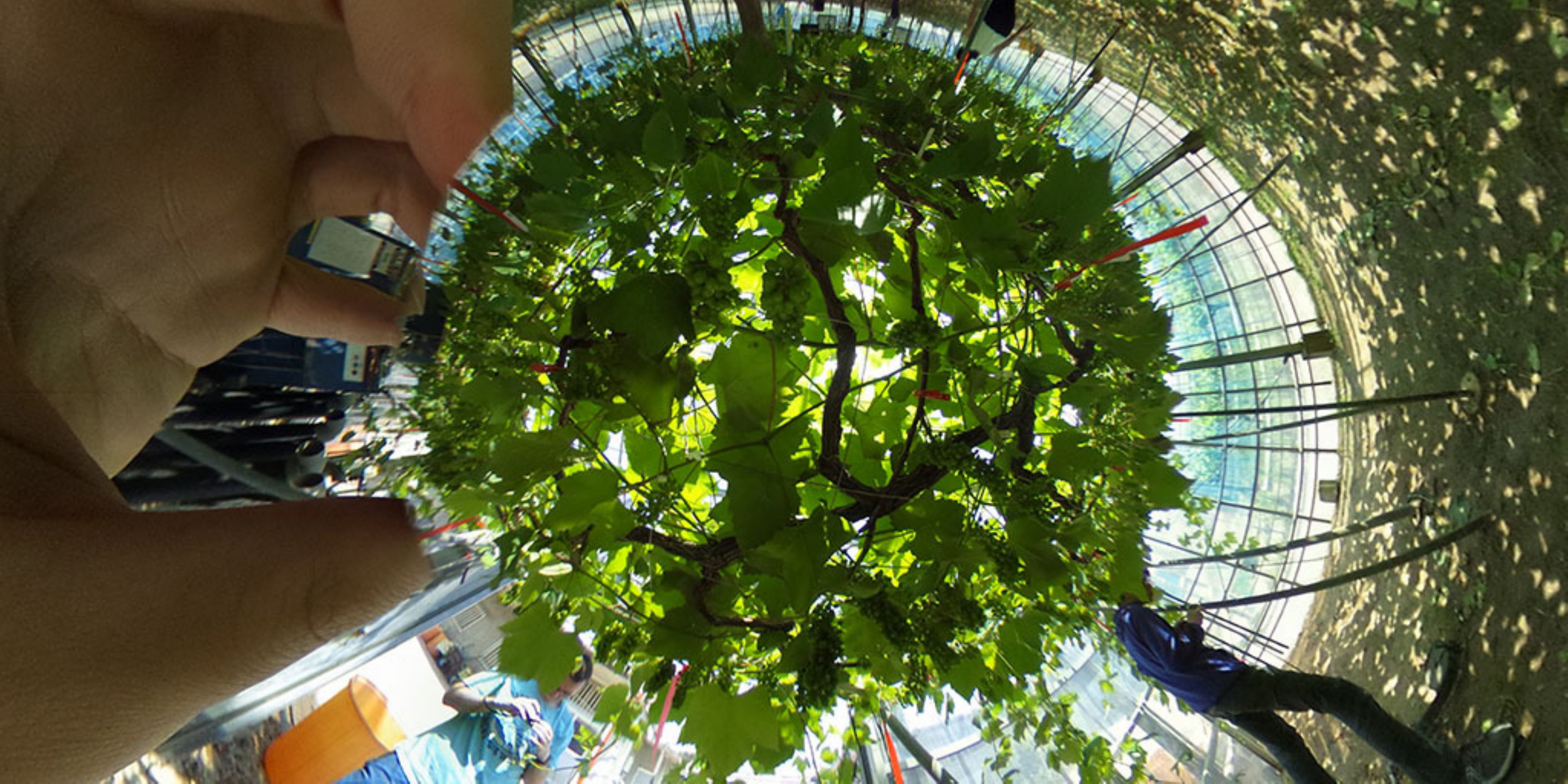}  
 \label{fig:trellis_stereo}
 }
 \caption{Examples of trellis images taken from below.} 
 \label{fig:trellis}
\end{figure*}
%
However, conventional object counting methods such as \cite{liCSRNetDilatedConvolutional2018,liuDecideNetCountingVarying2018,maBayesianLossCrowd2019,wanAdaptiveDensityMap2019,xiongOpenSetClosed2019, xieMicroscopyCellCounting2016,heAutomaticMicroscopicCell2019, guerrero-gomez-olmedoExtremelyOverlappingVehicle2015, mundhenkLargeContextualDataset2016, amatoCountingVehiclesDeep2019, Aich2017,Dobrescu2017,Giuffrida2018,Ubbens2018, Lempitsky2010} assume images are perspective images and are not suitable for omnidirectional images.
As shown in \figurename~\ref{fig:trellis}, omnidirectional images have particular distortion caused by projecting a spherical field of view onto a 2D plane.
In contrast to target objects in perspective images, the appearance of the target object changes according to location in omnidirectional images.
This distortion degrades the performance of conventional object counting methods.
One workaround is an iterative image conversion from an omnidirectional image to a perspective image~\cite{Iwamura_CHI2020_LBW}.
However, this is a complex process and requires special treatment for overlapping regions covered by multiple images.
Another workaround is to use a specially designed convolution for omnidirectional images such as \cite{Su2017,Cohen2018,Coors2018}.
However, it is known that such a convolution also degrades performance when compared with the standard convolution used on perspective images~\cite{Cohen2018,Coors2018}.


Therefore, we propose a new object counting method for omnidirectional images.
To the best of our knowledge, this is the first attempt to use omnidirectional images for object counting.
The proposed method uses omnidirectional images that have been transformed by a stereographic projection (called stereographic images hereafter), as shown in \figurename~\ref{fig:trellis_stereo}.
In contrast to equirectangular images, stereographic images are less distorted and the distortion is easier to manage.
In addition, we propose two methods to manage the remaining distortion.
One is a new data augmentation method designed for stereographic images, and 
the other is a distortion-adaptive Gaussian kernel that generates a density map ground truth taking into account the distortion of the stereographic projection.

Though we need an appropriate dataset to evaluate the proposed framework, to the best of our knowledge, there is no object counting dataset that consists of omnidirectional images.
Hence, we constructed an original grape-bunch dataset consisting of 527 omnidirectional images and their manually annotated ground truth.
We used this dataset to perform the experiments in this study.
The experimental results show that the mean absolute error (MAE) and mean squared error (MSE) of the proposed method are respectively 14.7\% and 10.5\% better than the results obtained by directly applying the conventional method.
\section{Related Work}
%
%
\subsection{Grape detection}
Grape detection methods can be modified for counting grapes. 
Grape detection is an especially important task because it is an essential technique for predicting yield and constructing automatic harvesting systems.
Thus, many grape detection methods have been proposed~\cite{Liu2015d,Tang2016,Seng2018,Valente2012,Xiong2018,Berenstein2010,Roscher2014,Nuske2015, Skrabanek2018,He2017,Santos2019a}.

Color is the simplest way to detect grapes.
Some methods use images taken in the daytime, and detect grapes with color-based classifiers~\cite{Liu2015d,Tang2016,Seng2018}.
Color-based detection methods are easy to implement but are easily affected by sunlight. 
To avoid the influence of sunlight, some methods use illuminated grape field images at night~\cite{Valente2012,Xiong2018}.
These methods are suitable for ripe grapes whose color is different from the color of the leaves.
However, it is difficult to detect green grapes that have the same color as the leaves.
Because we focus on counting green grape bunches, color-based methods are not suitable for our problem.

Using the shape of berries is another strategy for detecting grapes.
Shape-based methods can detect both ripe and green berries because they detect circles using edge information.
Methods using edge information~\cite{Berenstein2010} and the Hough transformation~\cite{Roscher2014,Nuske2015} have been proposed and 
obtained better detection results than color-based methods.
These methods can detect bunch regions but cannot recognize individual bunches.
Therefore, we still need to develop a bunch counting method to achieve the aim of this study.

Grape detection methods based on deep neural networks have also been proposed.
{\'{S}}krab{\'{a}}nek proposed a ConvNet-based grape region detection method~\cite{Skrabanek2018}.
This method achieves high detection accuracy but cannot detect bunches.
Santos et al. used Mask R-CNN~\cite{He2017} to detect grape bunches, and successfully detected instance-level grape bunches~\cite{Santos2019a}.
However, this method does not detect occluded bunches well.
As the image in \figurename~\ref{fig:trellis_stereo} shows, grape bunches in the images we use have occlusion and overlap.
Therefore, this method is not suitable for the bunch-counting problem.
\subsection{Object counting}
Grape bunch counting is a type of object counting task.
Object counting is a popular topic in computer vision research and has been studied for a long time.
People~\cite{liCSRNetDilatedConvolutional2018,liuDecideNetCountingVarying2018,maBayesianLossCrowd2019,wanAdaptiveDensityMap2019,xiongOpenSetClosed2019}, 
biological cells~\cite{xieMicroscopyCellCounting2016,heAutomaticMicroscopicCell2019}, 
cars~\cite{guerrero-gomez-olmedoExtremelyOverlappingVehicle2015, mundhenkLargeContextualDataset2016, amatoCountingVehiclesDeep2019}, 
and leaves~\cite{Aich2017,Dobrescu2017,Giuffrida2018,Ubbens2018}
are popular targets for the object counting. 
There is no grape bunch counting method as far as we know; therefore, we believe this paper is the first attempt to count grape bunches.

Object counting methods are roughly divided into detection-based and regression-based methods.
Detection-based methods use object detectors to determine the locations of object instances in an image  
and then count the detected instances~\cite{liDeepLearningBased2017,fernandez-gallegoWheatEarCounting2018,amatoCountingVehiclesDeep2019}.
However, most object detectors are not robust against occlusion, which degrades counting accuracy.
Thus, detection-based methods are inappropriate for grape-bunch counting because images for grape-bunch counting have a lot of occlusion and overlapping bunches.

Regression-based methods estimate the number of instances directly from image characteristics.
Usually, regression-based methods learn the dependency between dependent and independent variables.
However, this approach does not use the positions of instances, which helps count objects.
Therefore, Lempitsky et al.~\cite{Lempitsky2010} proposed a method that estimates the density map of the target object, and the number of instances is given by the integral of the density map.
Most recent object counting methods adopt the density map estimation method and obtain good  results~\cite{liCSRNetDilatedConvolutional2018,maBayesianLossCrowd2019,wanAdaptiveDensityMap2019,xiongOpenSetClosed2019,guerrero-gomez-olmedoExtremelyOverlappingVehicle2015, mundhenkLargeContextualDataset2016}.
Hence, we also apply a regression-based method that estimates a density map for counting grape bunches.
%
%
\subsection{Object detection for omnidirectional images}
An omnidirectional camera is a sensor whose field of view is the entire sphere.
Because omnidirectional cameras are able to capture all 360 degrees of a scene at once, they have been used 
for mobile robot navigation~\cite{Markovic2014,Jager2013,Yagi2001} and autonomous driving systems~\cite{DiasPais2019,Yahiaoui2019,Yogamani2019,Baek2018}.
Object detection methods using omnidirectional images were also proposed in \cite{Su2017, Cohen2018, Coors2018}.
 
The biggest problem with using omnidirectional images is distortion.
Object detection methods for omnidirectional images are categorized into two approaches according to how they deal with the distortion: by either using distorted images directly or by using special convolutions.
Using distorted images as they are is the simplest way to deal with omnidirectional images, and several methods that use this approach have been  proposed~\cite{DiasPais2019,Yahiaoui2019,Yogamani2019,Baek2018}.
However, because the appearance of an object changes according to its location in the image, the object detection accuracy of such a method is lower than that of a method using perspective images.

The special convolution-based approach considers the distortion and modifies it through the convolution. 
Methods such as SPHCONV~\cite{Su2017}, Spherical CNNs~\cite{Cohen2018}, SphereNet~\cite{Coors2018}, SpherePHD~\cite{Lee2019} take this approach.
These methods obtain better detection accuracy than the methods that use distorted images directly.
However, the integration of a unique convolution takes effort.

As described above, many object detection methods for omnidirectional images have been proposed, but there is no object counting method
using omnidirectional images as far as we know.
%
%
%
%
%
\section{Object Counting Method using Regression}
In this section, we present the framework for regression-based object counting methods for perspective images.
Then, we explain S-DCNet~\cite{xiongOpenSetClosed2019}, which we use for the proposed method.
\subsection{Framework}
We follow the regression-based object counting framework proposed in \cite{Lempitsky2010}, which adopts the most recent object counting methods.
In the framework, a regression model predicts a density map using an input image, and the number of objects is given by the integral of the density map.
The relationship between input images and density maps is learned from training data.

We define the density function, which gives us the density map ground truth.
Let $I_i$ be the $i$th training image, $p$ be a pixel in image $I_i$, $c(i)$ be the number of annotated objects in $I_i$, 
and $\boldsymbol{P}_i =\left\{P_i^{j}|j=1, 2, \ldots, c(i)\right\}$ be the positions of annotated objects in $I_i$.
The density function $ F^{0}_{i}(p)$ of the $i$th image at pixel $p$ is then given by
\begin{equation}
 F^{0}_{i}(p) = \sum_{j=1}^{c(i)}\mathcal{N}\left(p;P_{i}^{j},\sigma^2\boldsymbol{1}_{2 \times 2}\right),
  \label{eq:image_density}
\end{equation}
where $\mathcal{N}\left(p;P_{i}^{j},\sigma^2\boldsymbol{1}_{2 \times 2}\right)$ is a 2D Gaussian kernel with mean $P_{i}^{j}$ and isotropic covariance matrix $\boldsymbol{1}_{2 \times 2}$ with coefficient $\sigma$.
Usually, the value of $\sigma$ in \eqref{eq:image_density} is fixed to few pixels.
%
%
%
\subsection{S-DCNet}\label{sec:s-dcnet}
We use S-DCNet~\cite{xiongOpenSetClosed2019} for estimating the density map.
S-DCNet was originally proposed as a method for crowd counting.
As the range of the number of people extends to infinity, i.e., $[0, \infty)$, it is difficult to estimate the number of people from training data, which are usually finite.
S-DCNet solves this problem by dividing the input images and counting the people in sub-images so that their numbers are within the range desired for the training data.
S-DCNet has also been successfully applied to maize-tassel counting, which shows that S-DCNet can be effective for plant counting tasks.
Thus, we adopt it as the regression model of the proposed method.

In S-DCNet, the count value is discretized so that regression problem becomes a classification problem.
The discretization level should be adjusted for each problem.
\section{Proposed Method}
%
We propose a new object counting method for omnidirectional images. 
To make the conventional object counting methods suitable for omnidirectional images, we propose the use of stereographic images for object counting.
We also propose a new data augmentation method for stereographic images and a distortion-adaptive Gaussian kernel for managing the distortion.
We first explain stereographic projection and describe its properties.
Then, we describe both proposed methods.
\subsection{Stereographic projection}\label{sec:stereographic}
\begin{figure*}[tb]
 \centering
 \subfigure[Equirectangular projection]{
 \includegraphics[height=0.2\textwidth]{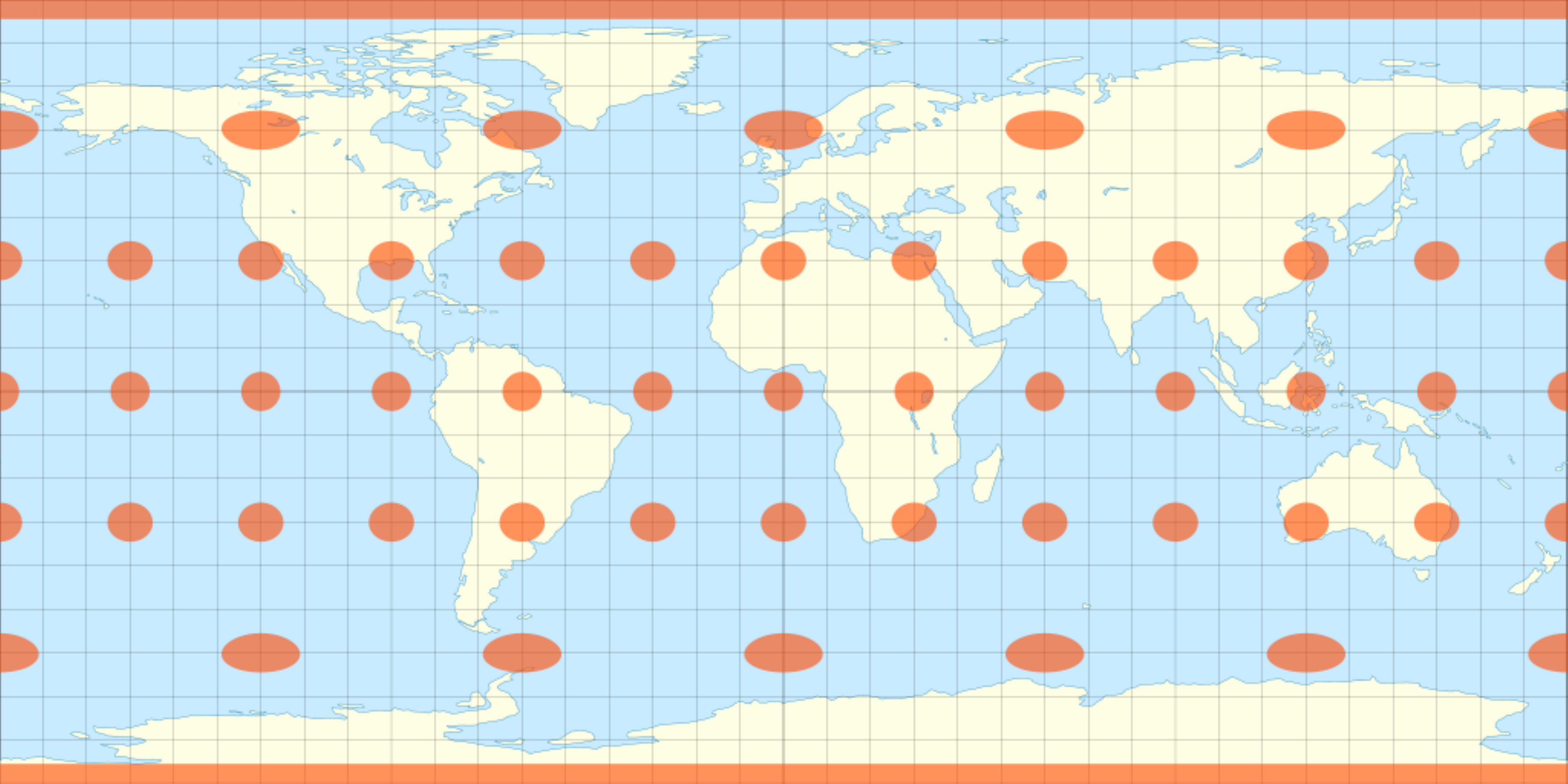}
 \label{fig:tissot_eq}
 }
 \subfigure[Stereographic projection]{
  \includegraphics[height=0.2\textwidth]{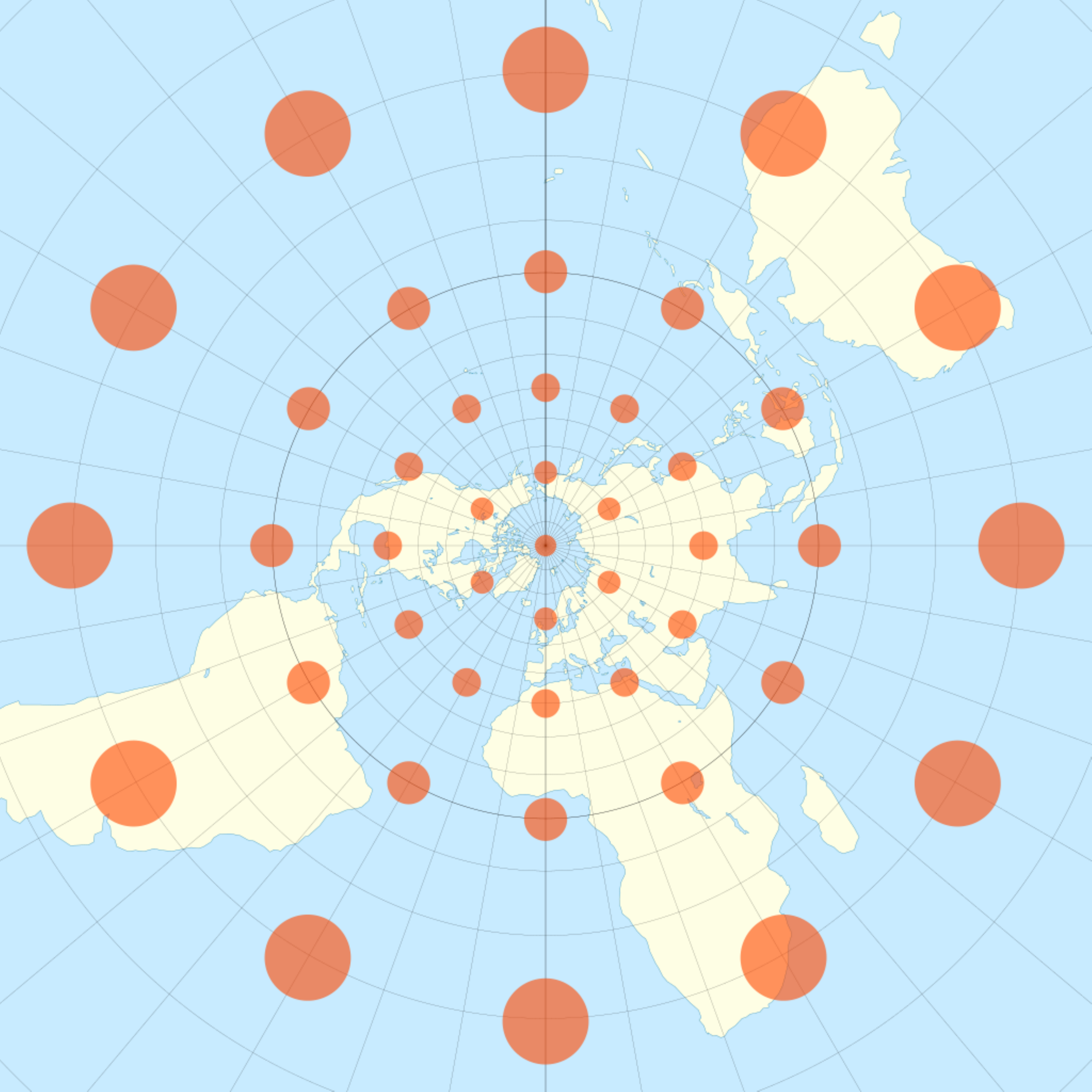}
 \label{fig:tissot_stereo} 
 }
  \caption{World map with Tissot's indicatrix (created by Justin Kunimune). The interval between the graticules is  $10^{\circ}$, and the diameter of the circles is 1,000 km.}
 \label{fig:tissot}
\end{figure*}
A stereographic projection~\cite{Snyder1987} maps a scene taken by an omnidirectional camera onto a 2D images, as shown in \figurename~\ref{fig:trellis_stereo}.
We give its formula and discuss its characteristics below.
%

We begin by presenting the projection model of omnidirectional images.
A scene taken by an omnidirectional camera is projected in two steps.
First, any points in 3D space are projected onto the unit sphere.
Next, the unit sphere is projected onto the image plane.

Of the various methods that project from the unit sphere to the plane, we use stereographic projection.
This is because images projected by stereographic projection have less distortion than those of equirectangular projection, which is the most popular projection method for omnidirectional images.
\figurename~\ref{fig:tissot} shows a world map with Tissot's indicatrix transformed by equirectangular and stereographic projections.
Tissot's indicatrix was proposed by Nicolas Auguste Tissot in 1859 to characterize the distortion of a projection from a sphere to a plane. 
As shown in \figurename~\ref{fig:tissot}, the circles on the equirectangular image are highly distorted.
Moreover, the circles on the stereographic image are not as distorted.

In addition, stereographic projection is the only projection in which small circles on the sphere are projected on the image plane as circles, although the centers of the projected circles are shifted slightly~\cite{doi:10.3138/carto.42.4.297}.
Therefore, circles on perspective images can be expressed as circles on stereographic images. 
Because a density map in the conventional framework for perspective images is expressed as a summation of 2D isotropic Gaussian kernels, which are circular, the density maps for stereographic images are also approximately expressed as 2D isotropic Gaussian kernels.

\begin{figure}[tb]
 \centering
 \includegraphics[width=0.4\textwidth]{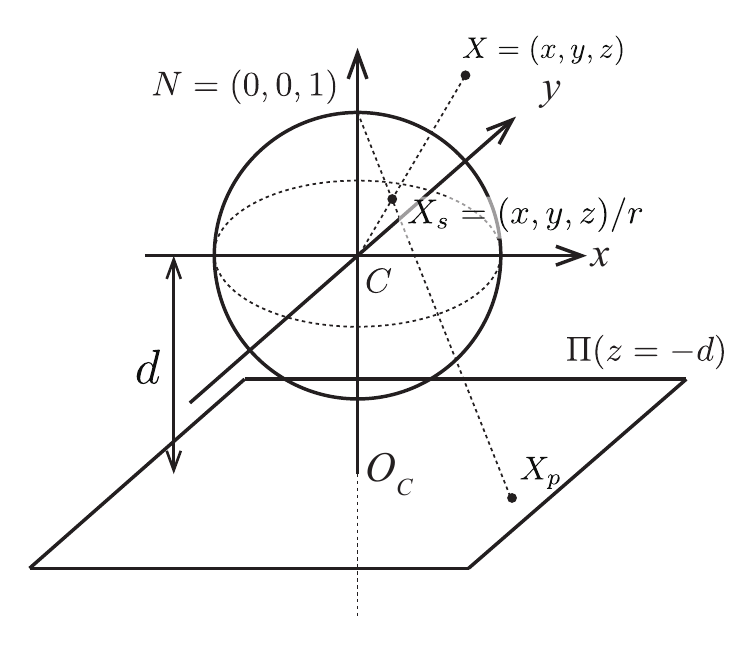}
 \caption{Model of stereographic projection.}
 \label{fig:stereographic}
\end{figure}

Let us formulate the stereographic projection.
As shown in \figurename~\ref{fig:stereographic}, we suppose a 3D space whose origin is the center of an omnidirectional camera $C$ and 
consider how the 3D space is projected onto plane $\Pi$ through spherical and stereographic projections.
First, any point $X=(x,y,z)$ of the 3D space is projected onto point $X_s$ on the unit sphere whose origin is $C$. 
Using $r = \sqrt{x^2+y^2+z^2}$, the 3D coordinate of $X_s$ is represented as
\begin{equation}
\label{eq:sphere_projection}
X_s = \left(\frac{x}{r},\frac{y}{r},\frac{z}{r}\right).
\end{equation}
Next, the surface of the unit sphere is projected onto a plane.
In general, stereographic projection is defined as a projection in which the unit sphere is projected onto a plane $z=0$ or tangent plane $z=-1$ of the south pole $S=(0,0,-1)$, with north pole $N=(0, 0, 1)$ as the projection center.
In this paper, expanding the original definition, we define stereographic projection as a projection in which the unit sphere is projected onto a plane $\Pi$ $(z=-d, \ d \geq 1)$.
Then, point $X_p$, which is the projection of $X_s$ onto plane $\Pi$, is given as
\begin{equation}
X_p = \left(\frac{1+d}{r-z}x, \frac{1+d}{r-z}y, -d\right).
\label{eq:plain_projection}
\end{equation}
%
Suppose plane $z = k$ is projected onto plane $\Pi$. 
Then, from (\ref{eq:plain_projection}), the coefficient of $x$- and $y$-coordinates, $(1+d)/(r-z)$, is a function of $r$. On plane $z=k$, $r$ is expressed as $r = \sqrt{x^2+y^2+k^2}$. 
Hence, coefficient $(1+d)/(r-z)$ is a function of $x^2+y^2$ on plane $z=k$.
This means that a plane in a stereographic image is distorted depending on the distance from the center of the image $O_c$, which is the intersection of $\Pi$ and the $z$ axis. 
%

\subsection{Data augmentation by image rotation}\label{sec:alignment}
Object counting problems such as crowd and plant counting have a shortage of annotated data, and 
most object counting methods solve this problem by data augmentation such as random cropping.
Because the distortion of perspective images is less than that of stereographic images, the distortion of randomly cropped images from perspective images 
is almost consistent.
However, stereographic images are distorted by the distance from the image center.
If random cropping is applied to stereographic images, the distortion of the cropped images will vary from image to image and the appearance of the object will also vary. 
Thus, it is inappropriate to use random cropping to augment stereographic image data.
%

\begin{figure}[tb]
 \centering
 \includegraphics[width=0.48\textwidth]{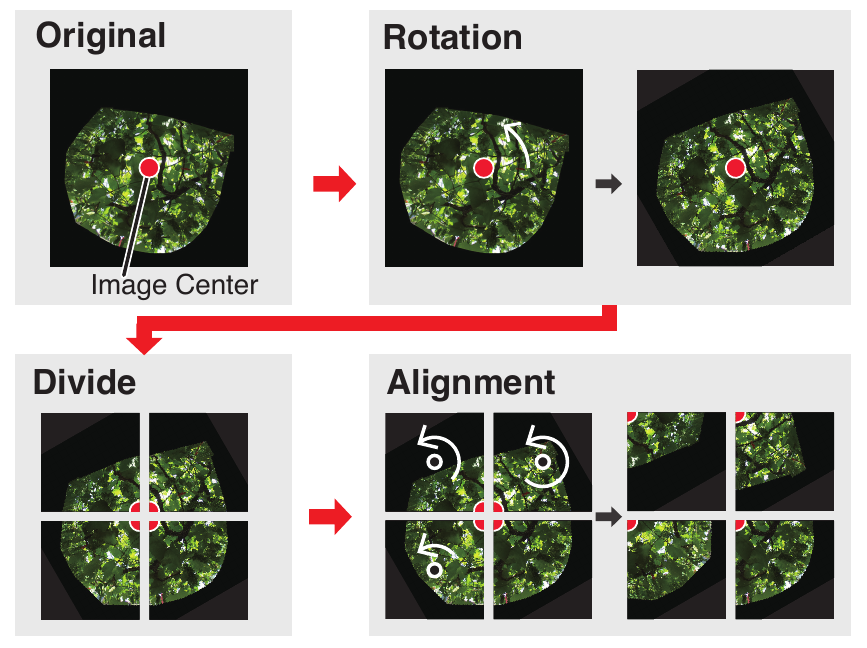}
 \caption{Proposed data augmentation and alignment method.}
 \label{fig:data_augmentation}
\end{figure}

Hence, we propose a data augmentation method for stereographic images.
The outline of the proposed method is shown in \figurename~\ref{fig:data_augmentation}.
We first rotate a stereographic image randomly about the image center.
Next, we divide the rotated stereographic image into four images.
Finally, we align the divided images by placing the image center in the upper-left corner.
Using this augmentation method, the distortions of the augmented images are the same, and the appearances of the objects in the same position are consistent.
%
%
%
%
\subsection{Distortion-adaptive Gaussian kernel}\label{sec:distortion_adaptive}
%
Because the scale of target objects in stereographic images varies because of distortion, the density function represented in~\eqref{eq:image_density} does not represent 
object density appropriately.
To overcome the multi-scale problem caused by the perspective distortion, various methods such as the maximum excess over sub arrays (MESA) distance for optimization~\cite{Lempitsky2010}, 
estimation of perspective distortion model~\cite{Shi2018}, 
learning-based ground truth estimation~\cite{Yan2019, Wan2019}, and 
multi-scale feature representation~\cite{Zhang2016a} have been proposed.
However, these methods do not consider the distortion of stereographic images.

\begin{figure}[tb]
 \centering
 \subfigure[Original image]{
 \includegraphics[width=0.22\textwidth]{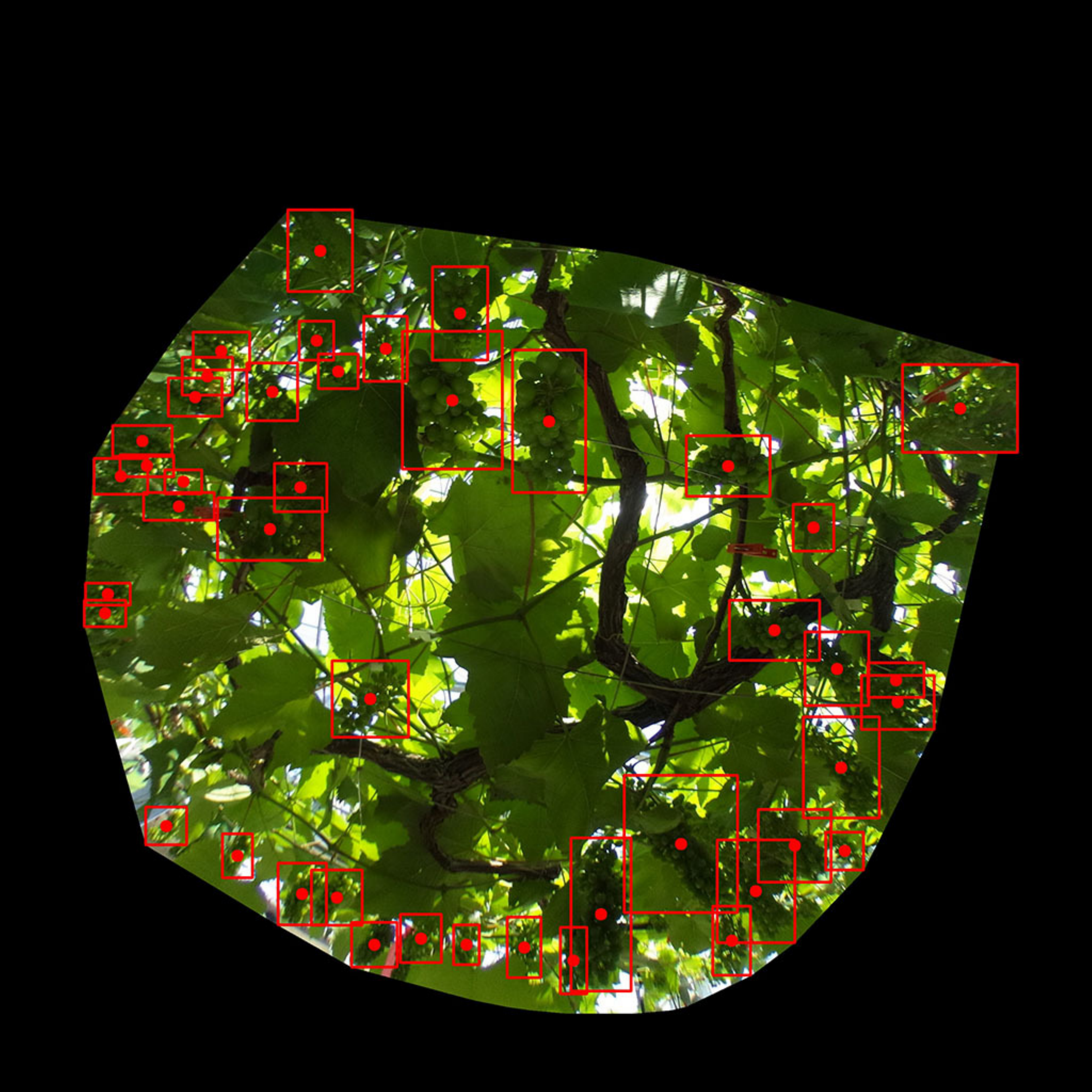}
 \label{fig:gt_original}
 }
 \subfigure[Fixed $\sigma$]{
 \includegraphics[width=0.22\textwidth]{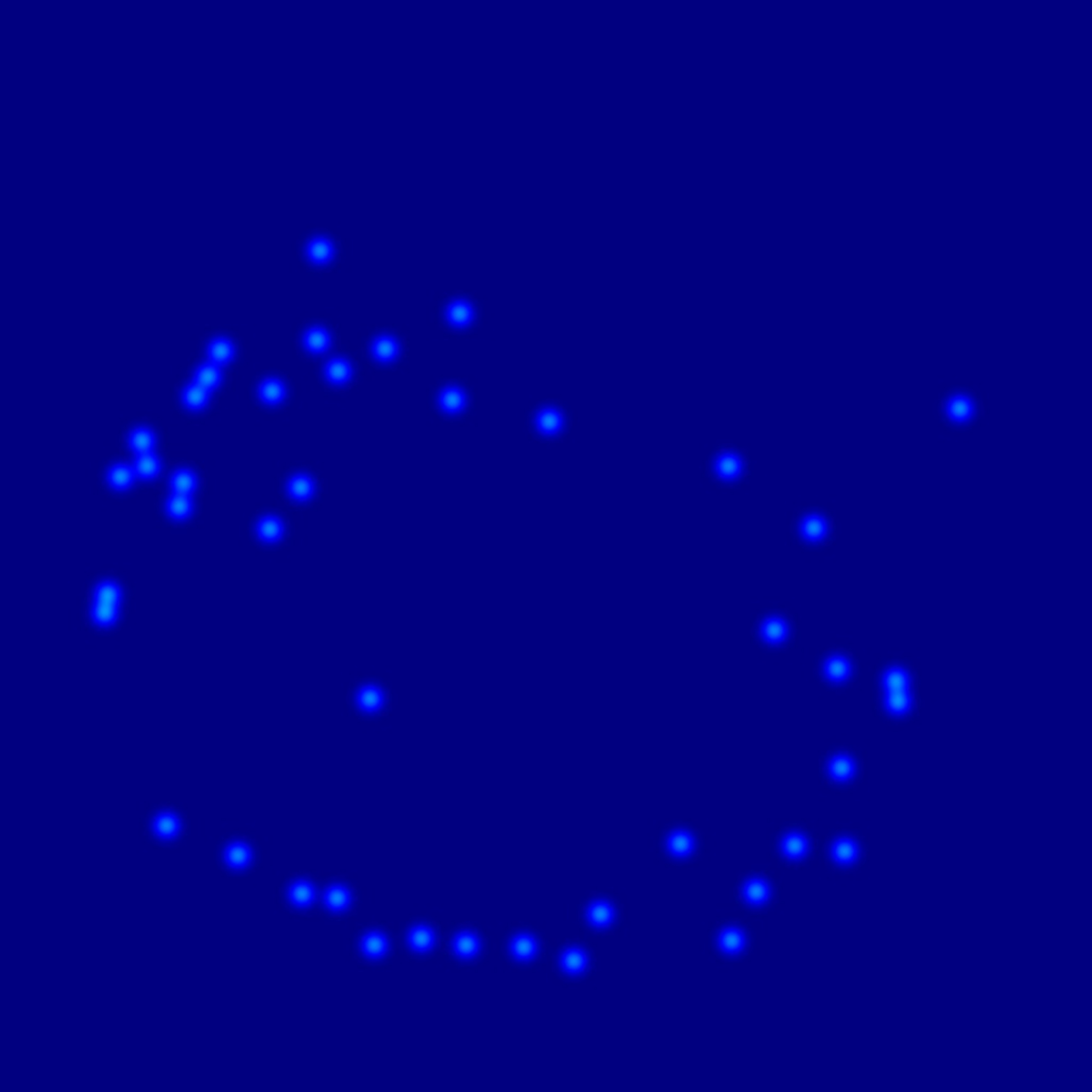}
\label{fig:gt_fix}
}
 \subfigure[Geometry-adaptive kernel~\cite{Zhang2016a}]{
 \includegraphics[width=0.22\textwidth]{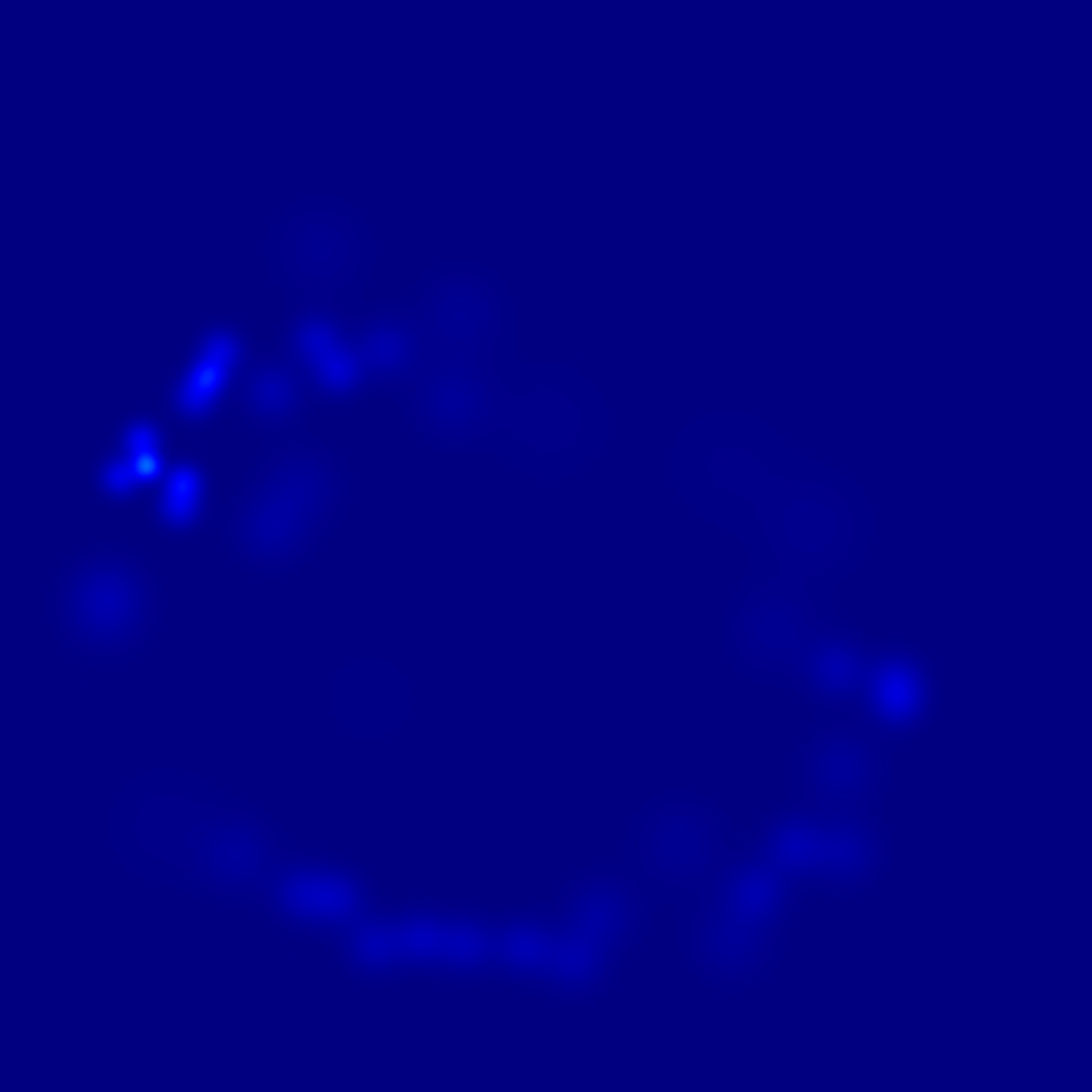}
\label{fig:gt_geo}
}
 \subfigure[Distortion-adaptive Gaussian kernel (proposed method)]{
 \includegraphics[width=0.22\textwidth]{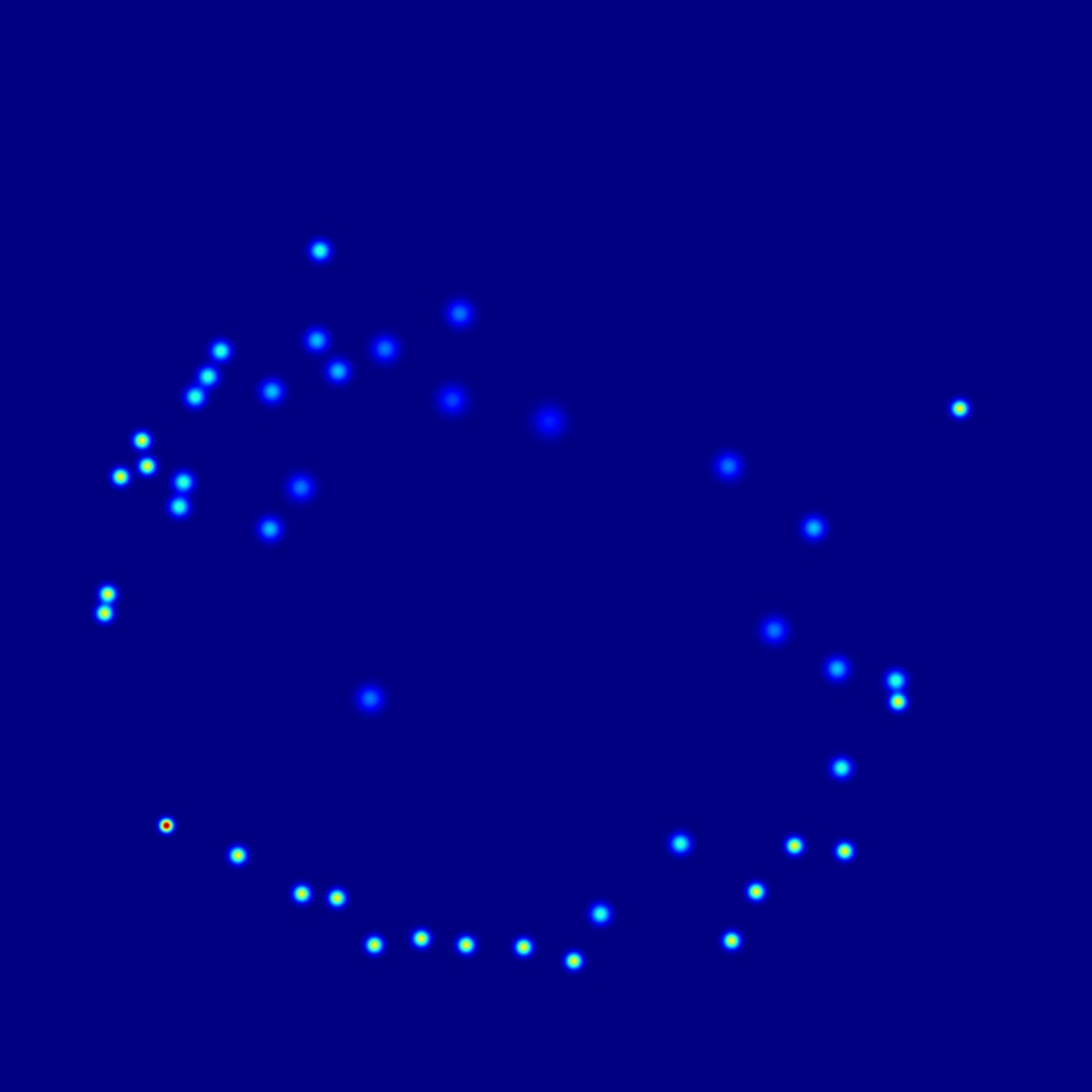}
 \label{fig:gt_proposed}
 }
 \caption{Density map ground truth generated by conventional methods and the proposed method. (a) Original image. Red rectangles and points are bounding boxes and their centers. Density map ground truth generated by (b) the fixed $\sigma$ shown in ~\eqref{eq:image_density}, , (c) geometry-adaptive kernel~\cite{Zhang2016a}, and (d) the proposed distortion-adaptive Gaussian kernel.}
 \label{fig:ground_truth}
\end{figure}

A model of the distortion of a stereographic projection image is given in Section~\ref{sec:stereographic}.
The change in scale is roughly in inverse proportion to the distance from the image center in ~\eqref{eq:plain_projection}.
Thus, we generate density map ground truth using \eqref{eq:image_density} with a 2D Gaussian kernel whose $\sigma$ is
in inverse proportion to the distance from the image center. 
We call this kernel the distortion-adaptive Gaussian kernel, and 
it is represented as follows.
Let $(u_i^{j}, v_i^{j}), (u_i^{c}, v_i^{c})$ be the coordinates of $P_i^{j}$ and the center of image $I_i$.
Then, the distance between $P_i^{j}$ and the image center $D(P_i^{j})$ is expressed as
\begin{equation}
 D(P_i^{j}) = \sqrt{\left(u-u_c\right)^2 + \left(v-v_c\right)^2}.
\end{equation}
The distortion-adaptive Gaussian kernel is then represented as   
\begin{equation}
    \mathcal{N}\left(p;P_{i}^{j},(\sigma(P_i^{j}))^2\boldsymbol{1}_{2 \times 2}\right),
\end{equation}
where coefficient $\sigma(P_i^{j})$ is defined with a constant $\sigma_\alpha$ as 
\begin{equation}
 \sigma(P_i^{j}) = \frac{1}{D(P)}\sigma_\alpha.
\label{eq:sigma}
\end{equation} 
The density function is expressed as \eqref{eq:image_density}, replacing $\mathcal{N}\left(p;P_{i}^{j},\sigma^2\boldsymbol{1}_{2 \times 2}\right)$ with the distortion-adaptive Gaussian kernel.
Examples of the density maps are shown in \figurename~\ref{fig:ground_truth}.
%

\section{Grape-bunch counting dataset}\label{sec:dataset}
We constructed a grape bunch dataset to train and evaluate the proposed method.
The dataset consists of 527 omnidirectional images.
The images were generated by stereographic projection, and the unit areas for counting are cropped.
The resolution of all dataset images is 2,688$\times$2,688 pixels, and 
bunches in the unit area are annotated by bounding boxes.
An example image from the dataset is shown in \figurename~\ref{fig:gt_original}.
The average number of bunches per image is 40.5.
Hereafter, we explain the data collection setup and the image processing used to construct the dataset.
\subsection{Data collection setup}\label{sec:data_collection}
We recorded images in the grape field of the Research Institute of Environment, Agriculture and Fisheries, Osaka Prefecture 
on May 13 and 20, 2019.
The weather of both recording days was fair.
The variety of grape is the Delaware and the grapes were grown on two trellises that are parallel to the ground in greenhouses.
The omnidirectional camera used for recording was the Theta S (Ricoh), which consists of two fisheye cameras and a gyroscope.

To follow the process used by farmers to count grape bunches described in Section~\ref{sec:intro}, we estimate the number of grape bunches in each 2 m $\times$ 2 m unit area.
Therefore, we divided the trellises into unit areas and recorded each unit area.
We put a marker on the corner of each unit area and attached an identification number to each area so we could recognize them. 
Each trellis had 60 unit areas.
We recorded images with the camera at the center of each unit area, which was estimated by eye, and 
at a distance of 0.5--1.5 m from the trellis.
We recorded two or three images for each unit area each day.
We took 268 images from one trellis and 259 images from the other trellis, which is a total of 527 images.
The raw images were 5,376$\times$2,688 pixels in JPG format and were stored as equirectangular images, as shown in \figurename~\ref{fig:trellis_eq}.
The Exif tag of the images includes the gyroscope data.
\subsection{Image processing}
To make the recorded images suitable for grape bunch counting, we preprocessed the raw images.
First, we converted the raw images to stereographic images.
We rotated the images on the unit sphere in the positive direction of the $z$ axis in \figurename~\ref{fig:stereographic}, which corresponds 
with the vertical downward direction when the images were recorded. 
The rotation angle and orientation were calculated based on the gyroscope data. 
An example of a stereographic projection image is shown in \figurename~\ref{fig:trellis_stereo}.

\begin{figure}[tb]
 \centering
 \includegraphics[width = 0.48\textwidth]{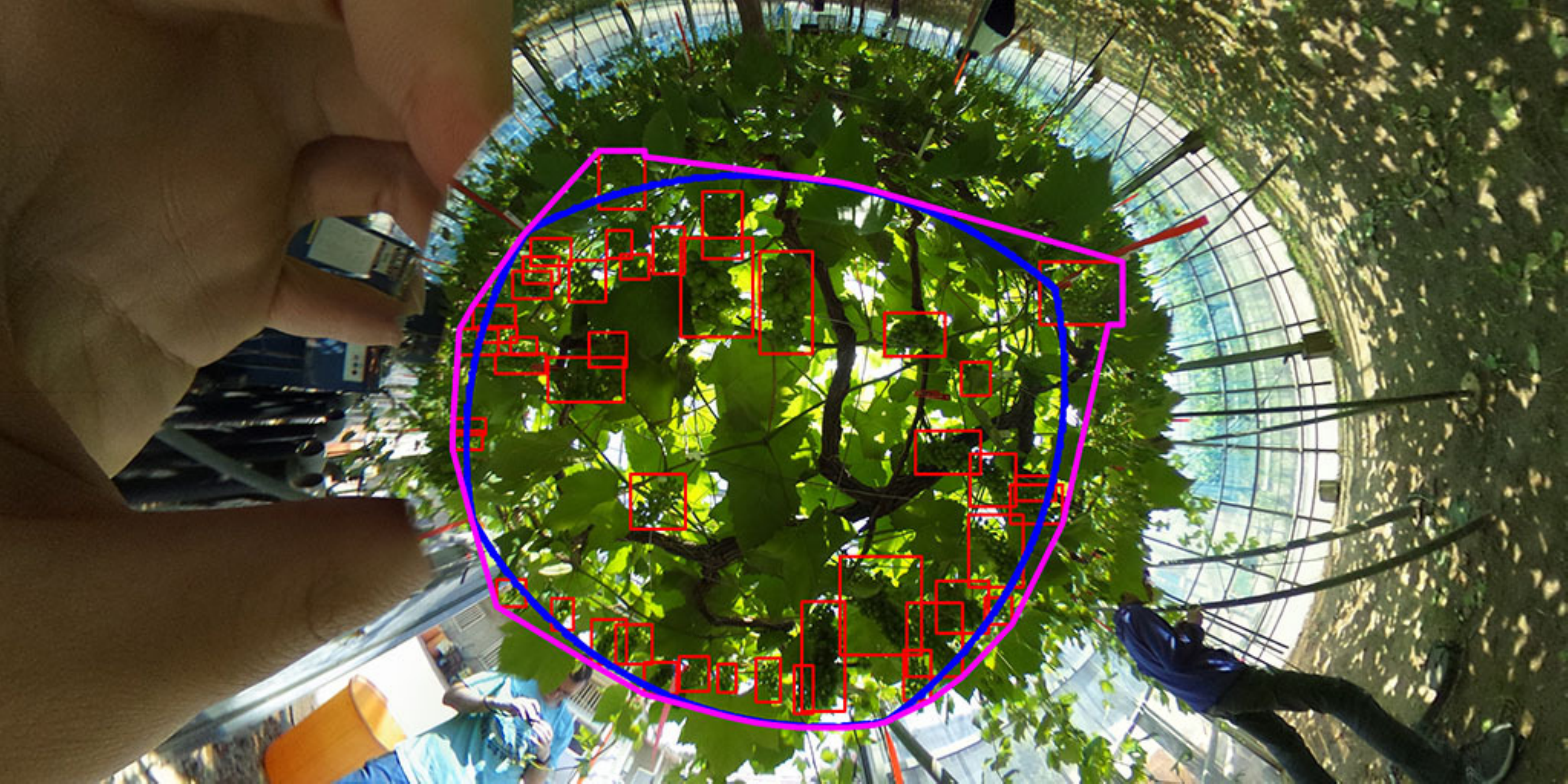}
 \caption{Construction of the dataset. Target region (blue line), bunch annotations (red rectangles), and cropped region (pink line).}
 \label{fig:db_line}
\end{figure}

Next, we cropped the 2 m $\times$ 2 m unit areas from the images.
As mentioned in Section~\ref{sec:data_collection}, there are markers on the corners of each unit area.
We first acquired the coordinates of the corners and drew a contour connecting the corners.
An example of this contour is shown in \figurename~\ref{fig:db_line} (blue line).
If we crop a unit area that is enclosed in this line, some grape bunches that lie across the line are cut off in the cropped image.
To avoid this, we first annotated the grape bunches inside the unit area and then cropped the unit area such that it includes these bunches.
As indicated by the red rectangles in \figurename~\ref{fig:db_line}, we annotated grape bunches inside the unit area with bounding boxes and then cropped the image using a convex hull of the union of the bounding boxes and the unit area enclosed by the lines.
The pink line in \figurename~\ref{fig:db_line} indicates the cropped region in the example.
\section{Experiments}
We evaluated the proposed methods using the dataset described in Section~\ref{sec:dataset}.
In this section, we explain the experimental setting and present the results.
\subsection{Experimental setting}
We divided the dataset into training and test sets according to trellis: 
the 269 images from the first trellis were used for training and the 259 images from the other trellis were used for testing.
The training data were augmented using the proposed augmentation method described in Section~\ref{sec:alignment}.
The rotation angle $\theta$ was given by a random number such that $0<\theta<\pi/2$.
We executed the data augmentation method twice on the training set.
We also divided the images of the training set into four images and used these divided sub-images for training.
In addition, we flipped the training set and executed the same process on the flipped data. 
Thus, the number of images for training was $268\times(1+2)\times4\times2=6,432$.
We reduced the resolution of the training set to $1,024\times1,024$ pixels and divided the images so that the 
resolution of the training data was $512\times512$ pixels.
When we generated the density map ground truth, the position of a bunch was given by the center of its bounding box and 
the constant $\sigma_\alpha$ in \eqref{eq:sigma} was fixed to either $12$ or $24$.

We conducted experiments with code implemented by the authors of \cite{xiongOpenSetClosed2019}~\footnote{https://github.com/xhp-hust-2018-2011/S-DCNet}.
We chose two discretization steps to adjust the grape bunch counting problem: 0.05 for the ranges from 0 to 0.5 and 0.5 for the range from 0.5 to $\infty$. 
Therefore, we discretized the count value as $\{0\}, (0,0.05], (0.05, 0.1], \ldots,  (0.45, 0.50], (0.5, 1.0], \ldots, \\ (C_{\mbox{max}}-0.5, C_{\mbox{max}}],(C_{\mbox{max}},\infty)$, where $C_{\mbox{max}}$ is the maximum number of bunches in the training data.
The default settings of the other parameters were used for training and evaluation.

To evaluate the proposed method, we conducted eight experiments while changing the settings of the augmented data alignment and 
image density ground truth.
We evaluated two conditions for the training data: divided images with and without alignment, and  
evaluated three conditions for generating image density ground truth: a fixed $\sigma$ in \eqref{eq:image_density}, the geometry-adaptive kernels proposed in~\cite{Zhang2016a}, and 
the proposed distortion-adaptive Gaussian kernels with $\sigma_\alpha = 12, 24$.
The geometry-adaptive kernel is a method that roughly estimates the perspective distortion using the distance between objects and adapts the 
$\sigma$ of the 2D Gaussian accordingly. 
In addition, the geometry-adaptive kernel is that used in the original S-DCNet.
The results were evaluated by MAE and MSE. 
\subsection{Results}
The experimental results are shown in TABLE~\ref{tab:results}.
When the geometry-adaptive kernel without alignment, which corresponds with S-DCNet, is used, it was directly applied to the stereographic images. 
When the distortion-adaptive Gaussian kernel with $\sigma_\alpha = 12$ and alignment are used, the best MAE and MSE are obtained, improving MAE by 0.59 and MSE by 0.54 with respect to the method in which S-DCNet is directly applied to stereographic images.
This shows that the proposed methods are effective for stereographic images.

In addition, the result using alignment data are better than those using unaligned data, regardless of which density function was used.
This shows that the alignment improves the accuracy of the object-counting results. 

The distortion-adaptive Gaussian kernel also improves accuracy.
For example, when compared with the density functions, the results obtained using the distortion-adaptive Gaussian kernel with $\sigma_\alpha = 12$ have the best MAE and MSE with and without alignment.

\begin{figure}
    \centering
    \subfigure[Input image. Red dots and boxes indicate the centers and bounding boxes of the bunches, respectively.]{
    \includegraphics[width = 0.42\textwidth]{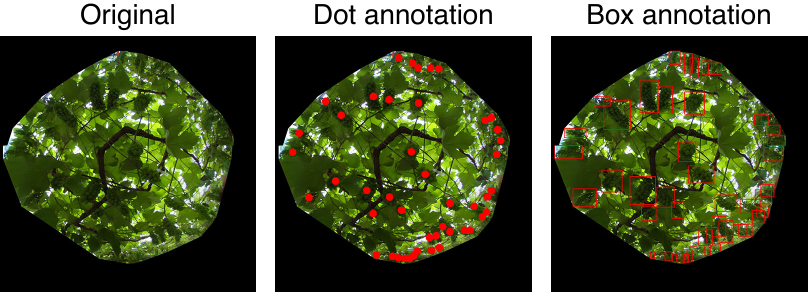}
    \label{fig:input}
    }
    \subfigure[Density map ground truth and estimated density maps. The lower right numbers in the density maps are the ground truth or the estimated number of bunches.]{
    \includegraphics[width = 0.42\textwidth]{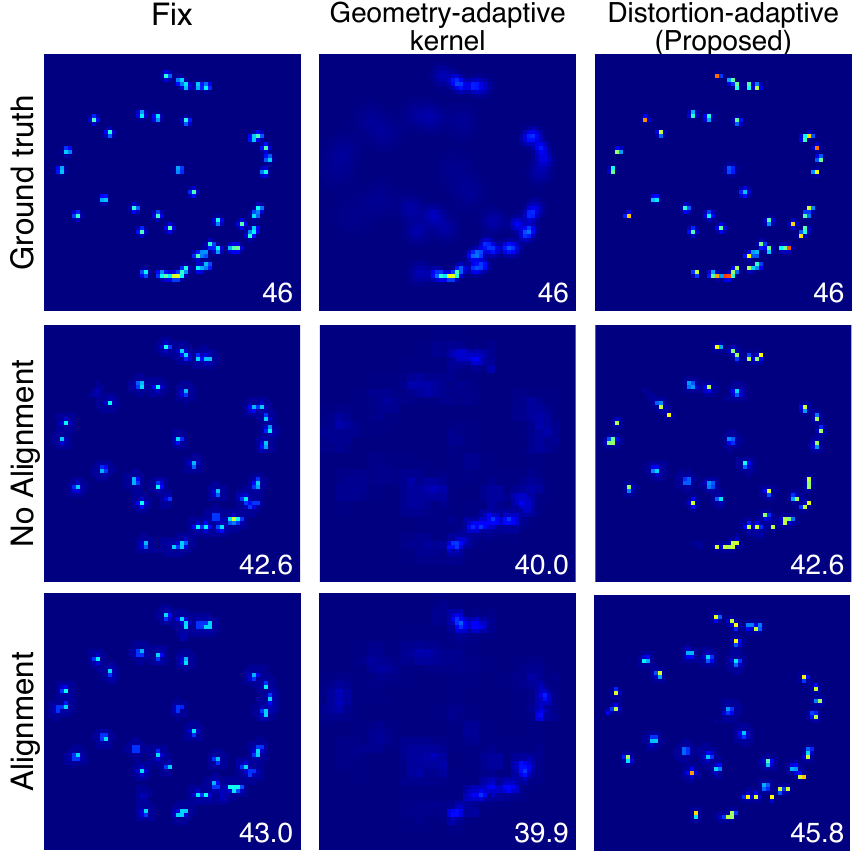}
    \label{fig:density_map}
    }   
    \caption{Input image and estimated density maps.}
    \label{fig:results_maps}
\end{figure}

An example of a test image, its ground truth, its estimated density maps, and the number of bunches is shown in \figurename~\ref{fig:results_maps}.
The input image shown in \figurename~\ref{fig:input} has 46 annotated bunches.
The sizes of the bunches are almost equal, but their scales in the image vary as a result of the distortion.
In the results in \figurename~\ref{fig:density_map}, the geometry-adaptive kernel performed the worst.
This kernel varies the kernel size according to the density of the objects, but  
it does not consider omnidirectional distortion.
However, the distortion-adaptive Gaussian kernel changes the size of the kernel based on the distortion.
Thus, the distortion-adaptive Gaussian kernel can obtain better estimation results on stereographic images.
In addition, the fixed-size and distortion-adaptive Gaussian kernels have the same estimation results without alignment, but the proposed method obtains better estimation results when alignment is used.
Other test images in which the proposed method obtained the best estimation results showed the same tendency.
This demonstrates that alignment improves accuracy and the distortion-adaptive Gaussian kernel boosts the effect of alignment.

%
\begin{table}[tb]
\centering
\caption{Experimental results. Stereographic images were used for the experiments. The bold values indicate the best results.}
 \begin{tabular}{l|c|c|c}
\hline
  Density function & Alignment & MAE & MSE\\
  \bhline
  \multirow{2}{*}{Fix ($\sigma = 8$)} & & 3.54 & 4.76\\ 
  \cline{2-4}
  & \checkmark & 3.47 & 4.72\\
  \hline
  \multirow{2}{*}{Geometry-adaptive kernel~\cite{Zhang2016a}} & & 3.99 & 5.12\\ 
  \cline{2-4}
  & \checkmark & 3.72 & 4.93\\
  \hline
  Distortion-adaptive ($\sigma_{\alpha} = 12$) &  & 3.46 & \textbf{4.58} \\
  \cline{2-4}
  (Proposed method)& \checkmark & \textbf{3.40} & \textbf{4.58}\\
  \hline
  Distortion-adaptive ($\sigma_{\alpha} = 24$) & & 5.41 & 6.59\\ 
  \cline{2-4}
  (Proposed method)& \checkmark & 5.18 & 6.24\\
  \hline
 \end{tabular}
 \label{tab:results}
\end{table}
%
%
\section{Conclusion}
In this paper, we proposed the use of omnidirectional images for object counting.
To easily deal with the distortion of omnidirectional images, we proposed the use of stereographic images, which are less distorted than equirectangular images, the most popular format of omnidirectional images.
However, because stereographic images are still distorted,  we proposed two methods to manage it. 
One is a data augmentation method: 
rotating the images around the image centers, dividing the rotated images into four sub-images, and placing the image centers in the upper-left corner.
As a result, the appearance of objects at a certain location in the augmented images becomes uniform, and training becomes easier.
The other method is a distortion-adaptive Gaussian kernel to generate the density map ground truth.
The kernel changes the coefficient $\sigma$ to be inversely proportional to the distance from the image center as the density map changes the distortion. 
We used grape bunches as a target object and conducted experiments on our original grape bunch dataset.
The results show that the proposed methods are effective for grape-bunch counting using omnidirectional images.
%
%
\section*{Acknowledgment}
We thank to JA Bank Osaka and the Telecommunications Advancement Foundation for supporting our research. 



\bibliographystyle{./IEEEtran}


\begin{thebibliography}{10}
\providecommand{\url}[1]{#1}
\csname url@samestyle\endcsname
\providecommand{\newblock}{\relax}
\providecommand{\bibinfo}[2]{#2}
\providecommand{\BIBentrySTDinterwordspacing}{\spaceskip=0pt\relax}
\providecommand{\BIBentryALTinterwordstretchfactor}{4}
\providecommand{\BIBentryALTinterwordspacing}{\spaceskip=\fontdimen2\font plus
\BIBentryALTinterwordstretchfactor\fontdimen3\font minus
  \fontdimen4\font\relax}
\providecommand{\BIBforeignlanguage}[2]{{%
\expandafter\ifx\csname l@#1\endcsname\relax
\typeout{** WARNING: IEEEtran.bst: No hyphenation pattern has been}%
\typeout{** loaded for the language `#1'. Using the pattern for}%
\typeout{** the default language instead.}%
\else
\language=\csname l@#1\endcsname
\fi
#2}}
\providecommand{\BIBdecl}{\relax}
\BIBdecl

\bibitem{liCSRNetDilatedConvolutional2018}
Y.~Li, X.~Zhang, and D.~Chen, ``{{CSRNet}}: {{Dilated Convolutional Neural
  Networks}} for {{Understanding}} the {{Highly Congested Scenes}},'' in
  \emph{Proc. of CVPR}, 2018, pp. 1091--1100.

\bibitem{liuDecideNetCountingVarying2018}
J.~Liu, C.~Gao, D.~Meng, and A.~G. Hauptmann, ``{{DecideNet}}: {{Counting
  Varying Density Crowds Through Attention Guided Detection}} and {{Density
  Estimation}},'' in \emph{Proc. of CVPR}, 2018, pp. 5197--5206.

\bibitem{maBayesianLossCrowd2019}
Z.~Ma, X.~Wei, X.~Hong, and Y.~Gong, ``Bayesian {{Loss}} for {{Crowd Count
  Estimation With Point Supervision}},'' in \emph{Proc. of ICCV}, 2019, pp.
  6142--6151.

\bibitem{wanAdaptiveDensityMap2019}
J.~Wan and A.~Chan, ``Adaptive {{Density Map Generation}} for {{Crowd
  Counting}},'' in \emph{Proc. of ICCV}, 2019, pp. 1130--1139.

\bibitem{xiongOpenSetClosed2019}
H.~Xiong, H.~Lu, C.~Liu, L.~Liu, Z.~Cao, and C.~Shen, ``From {{Open Set}} to
  {{Closed Set}}: {{Counting Objects}} by {{Spatial Divide}}-and-{{Conquer}},''
  in \emph{Proc. of ICCV}, 2019, pp. 8362--8371.

\bibitem{xieMicroscopyCellCounting2016}
W.~Xie, J.~Noble, and A.~Zisserman, ``Microscopy cell counting and detection
  with fully convolutional regression networks,'' \emph{Computer Methods in
  Biomechanics and Biomedical Engineering: Imaging \& Visualization}, pp.
  1--10, 2016.

\bibitem{heAutomaticMicroscopicCell2019}
S.~He, K.~T. Minn, L.~{Solnica-Krezel}, M.~Anastasio, and H.~Li, ``Automatic
  microscopic cell counting by use of deeply-supervised density regression
  model,'' in \emph{Medical {{Imaging}} 2019: {{Digital Pathology}}}, vol.
  10956, 2019, paper 109560L.

\bibitem{guerrero-gomez-olmedoExtremelyOverlappingVehicle2015}
R.~{Guerrero-G{\'o}mez-Olmedo}, B.~{Torre-Jim{\'e}nez}, R.~{L{\'o}pez-Sastre},
  S.~{Maldonado-Basc{\'o}n}, and D.~{O{\~n}oro-Rubio}, ``Extremely
  {{Overlapping Vehicle Counting}},'' in \emph{Proc. of Iberian Conference on
  Pattern Recognition and Image Analysis}, 2015, pp. 423--431.

\bibitem{mundhenkLargeContextualDataset2016}
T.~N. Mundhenk, G.~Konjevod, W.~A. Sakla, and K.~Boakye, ``A {{Large Contextual
  Dataset}} for {{Classification}}, {{Detection}} and {{Counting}} of {{Cars}}
  with {{Deep Learning}},'' in \emph{Proc. of ECCV}, 2016, pp. 785--800.

\bibitem{amatoCountingVehiclesDeep2019}
G.~Amato, L.~Ciampi, F.~Falchi, and C.~Gennaro, ``Counting {{Vehicles}} with
  {{Deep Learning}} in {{Onboard UAV Imagery}},'' in \emph{Proc. of 2019 {{IEEE
  Symposium}} on {{Computers}} and {{Communications}}}, 2019, pp. 1--6.

\bibitem{Aich2017}
S.~Aich and I.~Stavness, ``{Leaf Counting with Deep Convolutional and
  Deconvolutional Networks},'' in \emph{Proc. of ICCV 2017 Workshop on Computer
  Vision Problems in Plant Phenotyping}, 2017, pp. 2080--2089.

\bibitem{Dobrescu2017}
A.~Dobrescu, M.~V. Giuffrida, and S.~A. Tsaftaris, ``{Leveraging multiple
  datasets for deep leaf counting},'' in \emph{Proc. of 2017 ICCV Workshops},
  2017, pp. 2072--2079.

\bibitem{Giuffrida2018}
M.~V. Giuffrida, P.~Doerner, and S.~A. Tsaftaris, ``{Pheno-Deep Counter: a
  unified and versatile deep learning architecture for leaf counting},''
  \emph{Plant Journal}, vol.~96, no.~4, pp. 880--890, 2018.

\bibitem{Ubbens2018}
J.~Ubbens, M.~Cieslak, P.~Prusinkiewicz, and I.~Stavness, ``{The use of plant
  models in deep learning: An application to leaf counting in rosette
  plants},'' \emph{Plant Methods}, vol.~14, no.~1, pp. 1--10, 2018.

\bibitem{Lempitsky2010}
V.~Lempitsky and A.~Zisserman, ``{Learning To Count Objects in Images},'' in
  \emph{Proc. of NIPS}, 2010, pp. 1324--1332.

\bibitem{Iwamura_CHI2020_LBW}
M.~Iwamura, N.~Hirabayashi, Z.~Cheng, K.~Minatani, and K.~Kise, ``{VisPhoto}:
  Photography for people with visual impairment as post-production of
  omni-directional camera image,'' in \emph{Proc. CHI Extended Abstracts},
  2020.

\bibitem{Su2017}
Y.~C. Su and K.~Grauman, ``{Learning spherical convolution for fast features
  from 360$^{\circ}$ imagery},'' in \emph{Proc. of NIPS}, 2017, pp. 530--540.

\bibitem{Cohen2018}
T.~S. Cohen, M.~Geiger, J.~K{\"{o}}hler, and M.~Welling, ``{Spherical CNNs},''
  in \emph{Proc. of ICLR}, 2018, pp. 1--15.

\bibitem{Coors2018}
B.~Coors, P.~A. Condurache, and A.~Geiger, ``{SphereNet : Learning Spherical
  Representations for Detection and Classification in Omnidirectional
  Images},'' in \emph{Proc. of ECCV}, 2018, pp. 525--541.

\bibitem{Liu2015d}
S.~Liu and M.~Whitty, ``{Automatic grape bunch detection in vineyards with an
  SVM classifier},'' \emph{Journal of Applied Logic}, vol.~13, no.~4, pp.
  643--653, 2015.

\bibitem{Tang2016}
Y.~Tang, L.~Luo, C.~Wang, X.~Zou, W.~Feng, and P.~Zhang, ``{Robust Grape
  Cluster Detection in a Vineyard by Combining the AdaBoost Framework and
  Multiple Color Components},'' \emph{Sensors}, vol.~16, no.~12, paper 2098,
  2016.

\bibitem{Seng2018}
K.~P. Seng, L.~M. Ang, L.~M. Schmidtke, and S.~Y. Rogiers, ``{Computer vision
  and machine learning for viticulture technology},'' \emph{IEEE Access},
  vol.~6, pp. 67\,494--67\,510, 2018.

\bibitem{Valente2012}
A.~Valente, E.~Peres, J.~{Bulas Cruz}, O.~Contente, C.~Pereira, S.~Soares,
  R.~Morais, M.~Reis, P.~Ferreira, and J.~Baptista, ``{Automatic detection of
  bunches of grapes in natural environment from color images},'' \emph{Journal
  of Applied Logic}, vol.~10, no.~4, pp. 285--290, 2012.

\bibitem{Xiong2018}
J.~Xiong, Z.~Liu, R.~Lin, R.~Bu, Z.~He, Z.~Yang, and C.~Liang, ``{Green grape
  detection and picking-point calculation in a night-time natural environment
  using a charge-coupled device (CCD) vision sensor with artificial
  illumination},'' \emph{Sensors}, vol.~18, no.~4, paper 969, 2018.

\bibitem{Berenstein2010}
R.~Berenstein, O.~B. Shahar, A.~Shapiro, and Y.~Edan, ``{Grape clusters and
  foliage detection algorithms for autonomous selective vineyard sprayer},''
  \emph{Intelligent Service Robotics}, vol.~3, no.~4, pp. 233--243, 2010.

\bibitem{Roscher2014}
R.~Roscher, K.~Herzog, A.~Kunkel, A.~Kicherer, R.~T{\"{o}}pfer, and
  W.~F{\"{o}}rstner, ``{Automated image analysis framework for high-throughput
  determination of grapevine berry sizes using conditional random fields},''
  \emph{Computers and Electronics in Agriculture}, vol. 100, pp. 148--158,
  2014.

\bibitem{Nuske2015}
S.~Nuske, K.~Wilshusen, S.~Achar, L.~Yoder, S.~Narasimhan, and S.~Singh,
  ``{Automated Visual Yield Estimation in Vineyards},'' \emph{Journal of Field
  Robotics}, vol.~31, no.~5, pp. 837--860, 2014.

\bibitem{Skrabanek2018}
P.~{\'{S}}krab{\'{a}}nek, ``{DeepGrapes: Precise Detection of Grapes in
  Low-resolution Images},'' \emph{IFAC-PapersOnLine}, vol.~51, no.~6, pp.
  185--189, 2018.

\bibitem{He2017}
K.~He, G.~Gkioxari, P.~Dollar, and R.~Girshick, ``{Mask R-CNN},'' in
  \emph{Proc. of ICCV}, 2017, pp. 2980--2988.

\bibitem{Santos2019a}
T.~T. Santos, L.~L. de~Souza, A.~A. dos Santos, and S.~Avila, ``{Grape
  detection, segmentation and tracking using deep neural networks and
  three-dimensional association},'' \emph{Computers and Electronics in
  Agriculture}, vol. 170, no. 2020, paper 105247, 2019.

\bibitem{liDeepLearningBased2017}
W.~Li, H.~Fu, L.~Yu, and A.~Cracknell, ``Deep {{Learning Based Oil Palm Tree
  Detection}} and {{Counting}} for {{High}}-{{Resolution Remote Sensing
  Images}},'' \emph{Remote Sensing}, vol.~9, no.~1, paper 22, 2017.

\bibitem{fernandez-gallegoWheatEarCounting2018}
J.~{Fernandez-Gallego}, S.~Kefauver, N.~Aparicio, M.~Nieto-Taladriz, and
  J.~Araus, ``Wheat ear counting in-field conditions: {{High}} throughput and
  low-cost approach using {{RGB}} images,'' \emph{Plant Methods}, vol.~14,
  paper 22, 2018.

\bibitem{Markovic2014}
I.~Markovi{\'{c}}, F.~Chaumette, and I.~Petrovi{\'{c}}, ``{Moving object
  detection, Tracking and following using an omnidirectional camera on a mobile
  robot},'' in \emph{Proc. of ICRA}, 2014, pp. 5630--5635.

\bibitem{Jager2013}
B.~Jager, E.~Mair, C.~Brand, W.~Sturzl, and M.~Suppa, ``{Efficient navigation
  based on the Landmark-Tree map and the Z $\infty$ algorithm using an
  omnidirectional camera},'' in \emph{Proc. of IROS}, 2013, pp. 1930--1937.

\bibitem{Yagi2001}
Y.~Yagi, H.~Nagai, K.~Yamazawa, and M.~Yachida, ``{Reactive Visual Navigation
  Based on Omnidirectional Sensing ^^e2^^80^^93 Path Following and Collision
  Avoidance},'' \emph{Journal of Intelligent and Robotic Systems}, vol.~31, pp.
  379--395, 2001.

\bibitem{DiasPais2019}
G.~{Dias Pais}, T.~J. Dias, J.~C. Nascimento, and P.~Miraldo, ``{OmniDRL:
  Robust Pedestrian Detection using Deep Reinforcement Learning on
  Omnidirectional Cameras},'' in \emph{Proc. of ICRA}, 2019, pp. 4782--4789.

\bibitem{Yahiaoui2019}
M.~Yahiaoui, H.~Rashed, L.~Mariotti, G.~Sistu, I.~Clancy, L.~Yahiaoui, V.~R.
  Kumar, and S.~Yogamani, ``{FisheyeMODNet: Moving Object detection on
  Surround-view Cameras for Autonomous Driving},'' in \emph{Proc. of ICCV 2019
  Workshop on 360$^\circ$ Perception and Interaction}, 2019.

\bibitem{Yogamani2019}
S.~Yogamani, J.~Horgan, G.~Sistu, P.~Varley, D.~O. Dea, M.~U, S.~Milz,
  M.~Simon, K.~Amende, C.~Witt, H.~Rashed, S.~Chennupati, S.~Nayak, S.~Mansoor,
  X.~Perrotton, and P.~Patrick, ``{WoodScape : A multi-task , multi-camera
  fisheye dataset for autonomous driving c a},'' in \emph{Proc. of ICCV}, 2019,
  pp. 9308--9318.

\bibitem{Baek2018}
J.~Y. Baek, I.~V. Chelu, L.~Iordache, V.~Paunescu, H.~J. Ryu, A.~Ghiuta,
  A.~Petreanu, Y.~S. Soh, A.~Leica, and B.~M. Jeon, ``{Scene understanding
  networks for autonomous driving based on around view monitoring system},'' in
  \emph{Proc. of CVPR Workshops on Autonomous Driving}, 2018, pp. 1074--1081.

\bibitem{Lee2019}
Y.~Lee, J.~Jeong, J.~Yun, W.~Cho, and K.-j. Yoon, ``{SpherePHD: Applying CNNs
  on a Spherical PolyHeDron Representation of 360$^{\circ}$ Images},'' in
  \emph{Proc. of CVPR}, 2019, pp. 9173--9181.

\bibitem{Snyder1987}
J.~P. Snyder, \emph{{Map Projection: A Working Manual (U.S. Geological Survey
  Professional Paper 1395)}}.\hskip 1em plus 0.5em minus 0.4em\relax U.S.
  Government Printing Office, 1987.

\bibitem{doi:10.3138/carto.42.4.297}
D.~M. Goldberg and J.~R. Gott, ``Flexion and skewness in map projections of the
  earth,'' \emph{Cartographica: The Internatioanl Journal for Geographic
  Information and Geovisualization}, vol.~42, no.~4, pp. 297--318, 2007.

\bibitem{Shi2018}
M.~Shi, Z.~Yang, C.~Xu, Q.~Chen, M.~Shi, Z.~Yang, C.~Xu, Q.~Chen, P.-a. C.
  N.~N. For, C.~Counting, M.~Shi, Z.~Yang, C.~Xu, Q.~Chen, and S.~Member,
  ``{Perspective-Aware CNN For Crowd Counting},'' Inria Rennes - Bretagne
  Atlantique, Tech. Rep., 2018.

\bibitem{Yan2019}
Z.~Yan, Y.~Yuan, W.~Zuo, X.~Tan, Y.~Wang, S.~Wen, and E.~Ding,
  ``{Perspective-Guided Convolution Networks for Crowd Counting},'' in
  \emph{Proc. of ICCV}, 2019, pp. 952--961.

\bibitem{Wan2019}
J.~Wan and A.~Chan, ``{Adaptive Density Map Generation for Crowd Counting},''
  in \emph{Proc. of ICCV}, 2019, pp. 1130--1139.

\bibitem{Zhang2016a}
Y.~Zhang, D.~Zhou, S.~Chen, S.~Gao, and Y.~Ma, ``{Single-Image Crowd Counting
  via Multi-Column Convolutional Neural Network},'' in \emph{Proc. of CVPR},
  2016, pp. 589--597.

\end{thebibliography}
%



\end{document}